\documentclass[letterpaper]{article}
\usepackage{uai2020}
\usepackage[margin=1in]{geometry}

\usepackage{times}

\usepackage{url}
\usepackage{graphicx}
\usepackage{multirow}
\usepackage{amssymb,amsmath,amsthm}
\usepackage{subfig}
\usepackage{float}
\usepackage{booktabs}
\usepackage{footmisc}
\usepackage{algorithm}
\usepackage[noend]{algpseudocode}

\usepackage[backend=bibtex,bibstyle=numeric,citestyle=authoryear,maxnames=1]{biblatex}
\newcommand{\eat}[1]{}
\newcommand{\N}{\mathbb{N}}
\newcommand{\Z}{\mathbb{Z}}
\newcommand{\R}{\mathbb{R}}

\DeclareMathOperator{\sgn}{sign}
\newcommand{\poly}{\mathsf{poly}}
\newcommand{\NP}{\ensuremath\mathsf{NP}}
\newcommand{\Norm}[1]{\ensuremath\mathcal{N}\left(#1\right)}

\def\denseitems{
	\itemsep1pt plus1pt minus1pt
	\parsep0pt plus0pt
	\parskip0pt\topsep0pt}
\makeatletter
\newcommand{\algmargin}{\the\ALG@thistlm}
\makeatother
\newlength{\alginoutwidth}
\algnewcommand{\MultiState}[1]{\State%
  \parbox[t]{\dimexpr\linewidth-\algmargin}{\strut #1\strut}}
\algnewcommand{\InputState}[1]{
   \State \textbf{Input: } \parbox[t]{\dimexpr\linewidth-\algmargin-\alginoutwidth}{\strut #1 \strut}}
\algnewcommand{\OutputState}[1]{
   \State \textbf{Output: } \parbox[t]{\dimexpr\linewidth-\algmargin-\alginoutwidth}{\strut #1 \strut}}

\newtheorem{theorem}{Theorem}

\newtheorem{definition}{Definition}

\title{Verifying Individual Fairness in Machine Learning Models}
\author{{\bf Philips George John,~Deepak Vijaykeerthy,~Diptikalyan Saha\thanks{\{pgeorg04,deepakvij,diptsaha\}@in.ibm.com}}\\
IBM Research AI\\
Bengaluru 560~045, India\\
}

\begin{document}

\maketitle

\begin{abstract}
We consider the problem of whether a given decision model, working with structured data, has \emph{individual fairness}. Following the work of Dwork, a model is individually biased (or unfair) if there is a pair of valid inputs which are \emph{close} to each other (according to an appropriate metric) but are treated differently by the model (different class label, or large difference in output), and it is unbiased (or fair) if no such pair exists. Our objective is to construct verifiers for proving individual fairness of a given model, and we do so by considering appropriate relaxations of the problem. We construct verifiers which are sound but not complete for linear classifiers, and kernelized polynomial/radial basis function classifiers. We also report the experimental results of evaluating our proposed algorithms on publicly available datasets.
\end{abstract}
%\vskip .5pc
%Fourth Level Heading

\section{INTRODUCTION}\label{sec:introduction}

Recent breakthroughs in artificial intelligence, especially machine learning, have lead to AI-based systems assuming a significant role in making real-world decisions --- such as decision-making systems for recidivism risk assessments, credit assessments (including loan risk), hiring decisions, content dissemination in social media etc. Many of these systems are trained on, and then evaluate, \emph{structured data} pertaining to individuals (convicts, borrowers, job candidates etc). Unfortunately, studies have already shown that such systems may be prone to discriminating against users/consumers on the basis of characteristics such as race and gender \parencite{Angwin2016}, and this has even lead to legal mandates to ensure \emph{fairness} in such systems.

\eat{
The natural first question when considering the matter of fairness/bias in decision-making systems is to define what constitutes fairness/bias. Unfortunately, it is not possible to find a single definition of fairness which is appropriate for all situations. Definitions of fairness can be broadly grouped into two categories, namely, \emph{group fairness} (or \emph{statistical fairness}) and \emph{individual fairness}. A survey of the various definitions can be found in \parencite{Verma2018}. It is known that some of the common definitions of group and individual fairness are incompatible \parencite{Friedler2016,Kleinberg2017}, and as such, there is an inherent trade-off in choosing the fairness definition to use in a particular context.}

This paper tackles the problem of verifying the absence of individual bias in a given classifier (with \emph{white-box} access) which takes \emph{structured data} as input. \eat{The input is a vector of $n$ real numbers, each embodying a feature/attribute, some of which may be discrete (restricted to integer values). We assume that upper and lower bounds are known for each attribute, and that discrete attributes take contiguous integer values.} The definition of individual fairness/bias that we use in this paper is based on the abstract definition of individual fairness given by \parencite{Dwork2012}, which says that a model $f$ is fair if, for any pair of inputs $x, x^\prime$ which are sufficiently \emph{close} (as per an appropriate metric), the model outputs $f(x), f(x^\prime)$ are also close (as per another appropriate metric). Since using an $\ell_p$-metric for ``closeness'' does not take into account the structure of the data, we define a more flexible scheme which we feel is appropriate for our modality. We partition the input attributes into subsets, each of which is associated with an appropriate non-negative threshold, and say that two points $x, x^\prime$ are sufficiently close if the coordinate-wise absolute difference, $|x_i - x^\prime_i|$, for each attribute, is at most the threshold $\varepsilon_j$ corresponding to the subset $S_j$ to which the attribute index $i$ belongs. Then a model is \emph{fair} if, for any pair of close inputs, the model outputs are also close; for classification models, this means that the model decision does not change (i.e., $f(x) = f(x^\prime)$), whereas for regression models, this means that the absolute difference $|f(x) - f(x^\prime)|$ is sufficiently small. We note that this scheme is flexible enough to allow for (i) each attribute to have an independent threshold (as one extreme) including a threshold of zero for no perturbations, or (ii) the same threshold for all attributes (the other extreme), which is the same as using the $\ell_\infty$-metric for closeness (as is usual in adversarial robustness).

Another definition of individual fairness was used in~\parencite{Aggarwal2019,Galhotra2017,Udeshi2018} in the context of testing rather than verification. It is a simplified and non-probabilistic form of \emph{causal} or \emph{counterfactual fairness} \parencite{Kusner2017}, based on the notion of \emph{protected} or \emph{sensitive} attributes (in practical scenarios, these may be gender, race/ethnicity, religion etc.). The definition of \emph{fairness} in this context is that any two valid inputs which differ \emph{only on} the protected attribute(s) must always be put in the same class. Our definition subsumes this definition, by considering the threshold for the protected attributes to be sufficiently large (to allow arbitrary perturbations), and the threshold for the non-protected attributes to be zero (to disallow perturbations).

\emph{Challenges.} \eat{To the best of our knowledge, this is the first work that considers the problem of verifying individual fairness (as per Dwork) in machine learning models.} The challenges of verifying individual fairness, when compared to the existing work on verifying machine learning models, are two-fold: Firstly, the existing work on verifying bias/fairness in machine learning models considers notions of \emph{group fairness/bias} \parencite{Albarghouthi2017,Bastani2019}. An individual fairness property considers the worst case (fairness \emph{for all} similar input pairs, biased if \emph{there exists} a bad input pair), rather than the average case (with high probability, some notion of parity is maintained between different groups) considered in the group fairness definitions. Hence, \emph{the existing techniques for group fairness cannot be applied to verifying individual fairness}. Secondly, the other work on verification of ML models (which mostly considers the verification of \emph{adversarial robustness} -- see \parencite{Liu2019_Survey} for a survey) considers a \emph{local robustness} property; the verifier is given a \emph{nominal input} and it verifies robustness in the neighbourhood of that particular input (for example, given a particular image, the verifier either certifies that a small $\ell_\infty$-norm perturbation of that image does not change the class label, or provides a counter-example). However, verification of individual fairness notions requires us to check a \emph{global robustness} property (i.e. the classifier output does not change for perturbations of \emph{any} input in the domain). This means that \emph{existing approaches to local robustness verification are not directly applicable to our problem}.\\\\
\noindent{\bf {Our contributions.}} To the best of our knowledge, we present the first technique for individual fairness verification (global robustness) for ML models.

We give a meta-algorithm/framework for solving the verification problem, as well as particular algorithms for linear classifiers, and kernelized classifiers with polynomial/rbf kernels. Our algorithms are sound but incomplete (see section \ref{sec:meta_algo}), with the linear classifier case being an exception in that it is exact (both sound and complete) if we allow for worst-case exponential time.

%We also show that exact verification (both sound and complete) is $\NP$-Hard, even when all the attributes are continuous (relaxed case), for non-linear models.

\section{RELATED WORK}\label{sec:related_work}

In recent times, the software engineering community has addressed the problem of testing Individual fairness. THEMIS~\parencite{Galhotra2017} used random testing to generate test cases. AEQUITAS~\parencite{Udeshi2018} used random testing for global search and performs perturbation close to a sample which showed discrimination. \parencite{Aggarwal2019} used a combination of symbolic execution and model explainability techniques to systematically explore the decision space in a model instead of random testing. None of the above techniques guarantees absence of individual bias. 

Previous work on the verification of fairness in machine learning models has considered notions of group fairness such as disparate impact \parencite{Albarghouthi2017,Bastani2019}.
There have also been works considering the verification of the adversarial robustness property (and other similar properties) for machine learning models. Robustness does not have any notion of protected attributes. However, from an algorithmic perspective, our work is related to these, albeit using a different metric than the usual $\ell_p$-balls, and considering global robustness rather than local (as in the survey \parencite{ZhangHarman2019}).

These verification approaches can be broadly classified into (i) verification using tailor-made satisfiability modulo theory (SMT) or mixed integer linear programming (MILP) based approaches, and (ii) verification using convex relaxations. The former approach leads to sound and complete verification of certain classes of machine learning models --- such as linear models, decision trees (including tree ensembles), neural networks with piece-wise linear activation functions etc. \parencite{Katz2017,Ehlers2017,Bunel2018,Tjeng2019}, but at the cost of a worst-case exponential (or super-exponential) running time. The latter approach (convex relaxations) leads to efficient verifiers for properties such as (local) adversarial robustness for machine learning models \parencite{Kolter2017,Dvijotham2018,Raghunathan2018,Gowal2018,Singh2018,Wang2018a,Wang2018b,Wang2018c,Zhang2018b,Gehr2018,Mirman2018,Qin2019,Fazlyab2019,Salman2019,Singh2019}, but at the cost of sacrificing completeness. Another advantage is that such techniques can also be used for non-linear models, even though much of the existing work focuses on linear and piece-wise linear models.

Perhaps the work most closely related to ours (from a technical perspective) is that of \parencite{Raghunathan2018}, who verify the (local) adversarial robustness of a given feed-forward neural network with ReLU activations by relaxing adversarial robustness to a polynomial optimization problem and then using semidefinite relaxations to give lower bounds.
\section{PRELIMINARIES}\label{sec:prelims}

We consider models with a known \emph{prediction function} $f: \R^n \rightarrow \R$. We assume that we have white-box access to $f$; i.e. we have access to all the parameters and hyper-parameters which together give a closed-form expression for $f$. A regression model uses the output of $f$ directly as the predicted value of the regression variable. A binary classifier $h: \R^n \rightarrow \{\pm 1\}$ is of the form $h(x) = \sgn(f(x))$. We will often refer to the prediction function $f$ as the classifier itself, with the understanding that the predicted label for $x$ will actually be the sign of $f(x)$. We assume that $f$ is \emph{smooth}, or at least continuously differentiable twice ($C^2$).

The model $f$ takes as input a real vector with $n$ features, where the domain of feature $x_i$ is $\mathsf{Dom}_i := \{x \in \R \mbox{ or } \Z \mid l_i \leq x \leq u_i \}$. That is, each feature can be either continuous (in $\R$) or discrete (in $\Z$), and takes values in a fixed interval $[l_i, u_i]$. This characterization of input features is suitable for us since, in this work, we are chiefly considering decision models (classification/regression) on structured data. We say that an input sample is \emph{valid} if the domain constraints for all features are satisfied.

If $f: \R^n \rightarrow \R$ is a decision model, the abstract definition of individual fairness, given by \parencite{Dwork2012}, is as follows: Given appropriate distance functions --- $d(\cdot,\cdot)$ on $\R^n$ (the domain of $f$) and $D(\cdot,\cdot)$ on $\R$ (the co-domain of $f$) --- as well as thresholds $\varepsilon \geq 0$ and $\delta \geq 0$, the model is individually fair if, for any pair of inputs $x, x^\prime$ such that $d(x, x^\prime) \leq \varepsilon$, we have $D(f(x), f(x^\prime)) \leq \delta$. 

The intuition behind this notion of individual fairness is that \emph{small} or \emph{non-significant} perturbations of a sample $x$ to $x^\prime$ (i.e. the perturbations where $d(x, x^\prime) \leq \varepsilon$) must not be treated ``differently'' by a fair model. The choice of the input distance function $d(\cdot, \cdot)$ identifies the perturbations to be considered non-significant, while the choice of the output distance function $D(\cdot,\cdot)$ limits the changes allowed to the perturbed output in a fair model.

For classification models $f: \R^n \rightarrow [k]$, it is appropriate to use the discrete metric $D(y, y^\prime) := \mathbb{I}[y = y^\prime]$ with the threshold $\delta = 0$ on the model output since, in a fair classification model, we would want to prevent any change in the class label due to small perturbations of the input. For regression models, a simple choice would be the absolute error $D(y, y^\prime) := |y - y^\prime|$, with the threshold $\delta > 0$ chosen appropriately based on the scale of the regression variable.

Our notion of closeness in the input domain must take into account the structure of the data, and be general enough to give non-trivial and useful results (fairness certificates/bias instances) for a variety of structured datasets and models. So we proceed as follows. Let the input features be indexed as $[n] := \{1, \ldots, n\}$. We partition $[n]$ into disjoint sets $S_1, \ldots, S_t$, with corresponding thresholds $\varepsilon_1, \ldots, \varepsilon_t \geq 0$ chosen based on domain-specific knowledge of the dataset. A perturbation of $x \in \R^n$ to $x^\prime$ is considered non-significant if for all $j \in [t]$, and for all $i \in S_j$, we have $|x_i - x^\prime_i| \leq \varepsilon_j$.

\emph{Note: } For notational convenience, we allow the thresholds $\varepsilon_j$ to take a special value $\infty$, which implies that $x_i$ and $x^\prime_i$ can differ arbitrarily for all $i \in S_j$. From a algorithmic perspective, the constraint on $|x_i - x^\prime_i|$ would be removed for all such indices $i$. Also, if $\varepsilon_j = 0$ for any $j \in [t]$, we can eliminate the variables $x^\prime_i$ for all $i \in S_j$ to reduce the dimensionality of the problem.

We formally define the individual bias of a decision model as follows:
\begin{definition}[Individual bias]
A model $f: \R^n \rightarrow \R$ is said to be \emph{individually biased} if there exists a pair of valid inputs $x$ and $x^\prime$, with $|f(x) - f(x^\prime)| > \delta$, such that $|x_i - x^\prime_i| \leq \varepsilon_j$ for all $i \in S_j$, and for all $j = 1, \ldots, t$. Such a pair $(x, x^\prime)$ is called an individual bias instance of the model $f$.
\end{definition}

A \emph{linear (binary) classifier} is $h(x) = \sgn(f(x))$, where $f$ is of the form $f(x) = w^\top x + b$ for some $w \in \R^n$ and $b \in \R$. The decision boundary of a linear classifier is an affine hyperplane. A linear regression model is just $f(x) = w^\top x + b$, without the sign function.

A \emph{kernelized (binary) classifier} is $h(x) = \sgn(f(x))$, where $f$ is of the form $f(x) = \sum_{i = 1}^{M} w_i y_i\,K(x_i, x)$,
$S := \{(x_i, y_i) : i = 1, \ldots, M\}$ is a subset of the training set, $w_i$ is the weight assigned to the sample $(x_i, y_i) \in S$, and $K(\cdot,\cdot)$ is the Kernel function.

One of the commonly used kernels is the degree-$d$ polynomial kernel, where $K(x, y) = (a x^\top y + b)^d$ for some $a, b \in \R$. The prediction function of such a classifier $f(x) = \sum_{i = 1}^{M} w_i y_i (a x_i^\top x + b)^d$ can then be written as a degree-$d$ polynomial in the variables $x_1, \ldots, x_n$. A polynomial kernel allows for a ``curved'' non-linear decision boundary rather than the ``straight'' hyperplane boundary of a linear classifier.

Another commonly used kernel function is the radial basis function (RBF, also known as gaussian) kernel, with $K(x, y) = exp(-\gamma \cdot \|x - y\|^2)$. In an RBF kernelized classifier, high confidence negative decision regions can be seen around the negative data points, whereas high confidence positive decision regions can be seen around the positive data points. %The prediction function of such a classifier can be viewed as a combination of signed $n$-dimensional gaussians (with variances dependent on the weights $w_i$) centered around each of the points $x_i$. Intuitively, a kernelized RBF classifier with prediction function $f(x) = \sum_{i = 1}^{M} w_i y_i exp(-\gamma \cdot \|x_i - x\|^2)$ predicts the appropriate class $y_i$ with high confidence for data points which are close to one of the $x_i$-s, and predicts (with lower confidence) the classes of other data points based on the $x_i$-s which it is closer to. 

%\begin{figure}[t]
%	\centering
%	\subfloat[t][Linear]{
%      \includegraphics[width=0.45\columnwidth]{images/lin_boundary.png}
%	  \label{fig::prelims::dec_boundaries::lin}
%	}%
%	\subfloat[t][Polynomial]{
%	  \includegraphics[width=0.45\columnwidth]{images/poly_boundary.png}
%	  \label{fig::prelims::dec_boundaries::poly}
%	}\\%
%	\subfloat[t][RBF]{
%	  \includegraphics[width=0.45\columnwidth]{images/rbf_boundary.png}
%	  \label{fig::prelims::dec_boundaries::rbf}
%	}%
%	\caption{Decision boundaries of various classifiers.}\label{fig::prelims::dec_boundaries}
%\end{figure}

A set $S \subseteq \R^n$ is said to be a \emph{convex set} if for any two points $x, y \in S$ and any $0 \leq \lambda \leq 1$, the point $\lambda x + (1 - \lambda) y \in S$. If $S$ is a convex set, a function $f: S \rightarrow \R$ is said to be a \emph{convex function} if for any $x, y \in S$ and $0 \leq \lambda \leq 1$, $f(\lambda x + (1 - \lambda)y) \leq \lambda f(x) + (1 - \lambda) f(y)$. On the other hand, a function is said to be \emph{concave} if we have the inequality $f(\lambda x + (1 - \lambda)y) \geq \lambda f(x) + (1 - \lambda) f(y)$ for all $x, y \in S$ and $0 \leq \lambda \leq 1$.

%The importance of the notion of convexity in machine learning stems from the fact that it is possible to get tractable guarantees for solving convex optimization problems: \emph{minimizing} a convex function $f$ over a convex set $S$, which is generally not the case for non-convex problems. Similarly, we have tractable guarantees \emph{maximizing} a concave function $f$ over a convex set $S$. When we look at constraints, the feasible set of a constraint $h(x) \leq 0$ is convex (i.e. ``good'') when $h$ is a convex function, and the feasible set of a constraint $h(x) \geq 0$ is convex when $h$ is a concave function. The feasible set of a constraint $h(x) = 0$ (or equivalently, the combination of the two constraints $h(x) \geq 0$ and $h(x) \leq 0$) is convex/good only when $h$ is both convex \emph{and} concave, which happens only when $h$ is an \emph{affine} function, i.e., $h(x) = w^\top x + b$. This also provides some theoretical insight into why the verification problem is easier for linear models (where the prediction function $f$ is affine), which we shall see later.

\paragraph{Notation} $[n]$ denotes the set $\{1,2,\ldots,n\}$. If $x \in \R^n$, $x_{-i}$ denotes the tuple $(x_j)_{j \neq i} \in \R^{n-1}$ and for $S \subseteq [n]$, $x_S$ is shorthand for the tuple $(x_i)_{i \in S}$. If $S$ is a set, then $\binom{S}{k}$ denotes the set of all subsets $T \subseteq S$ with $|T| = k$. If $|S| = n$, then $\left| \binom{S}{k} \right| = \binom{n}{k} = \Theta(n^k)$. We use $\R^{m \times n}$ to denote the space of all $m \times n$ real matrices, and $\mathbb{S}^{n}(\R)$ to denote the space of all real symmetric $n \times n$ matrices. If $f$ is a function, $f \succeq 0$ denotes that $f$ is \emph{non-negative} or positive semi-definite (p.s.d); that is, $f(x) \geq 0$ for all $x$ in the domain of $f$. If $A$ is a matrix, then $A \succeq 0$ denotes that the matrix is positive semi-definite (p.s.d). We use $\N_d^n := \{ \alpha \in \N^n \mid |\alpha| := \alpha_1 + \ldots + \alpha_n \leq d\}$ to denote the set of multi-indices corresponding to the exponents of $n$-variate monomials with degree $\leq d$.
\section{VERIFYING INDIVIDUAL BIAS}
\subsection{A META-ALGORITHM}\label{sec:meta_algo}

Ideally, a \emph{verifier} for the individual bias property would solve the following decision problem (after fixing the attribute domains $\mathsf{Dom}_1, \ldots, \mathsf{Dom}_n$, the feature partitions $S_1, \ldots, S_t$, and the thresholds $\varepsilon_1, \ldots, \varepsilon_t$ and $\delta$): ``For a given model $f$, does there exist an individual bias instance?'' (\textsc{Yes/No}). This problem would be \emph{undecidable}, by Rice's theorem \parencite{Hopcroft1990Book}, if we allow $f$ to be an arbitrary partial recursive function. Even if we restrict the choice of $f$ to common non-linear function classes, the problem would still likely be $\NP$-hard (since we show that it is $\NP$-hard when $f$ is a polynomial with $\deg(f) \geq 2$).

{We focus on the following guarantees for the verifier $\mathcal{V}$ (with our desired property for the model $f$ being individual fairness/no bias):
\setlength{\parskip}{0em}
\begin{enumerate}\itemsep0em
\item \textbf{Soundness: } If the verifier $\mathcal{V}$ outputs \textsc{No Bias}, then the model $f$ is actually unbiased.
\item \textbf{Completeness: } If the verifier $\mathcal{V}$ outputs a bias instance, then the model $f$ is actually biased. Also, the verifier $\mathcal{V}$ will \emph{always} terminate with either \textsc{No Bias} or a valid bias instance.
\end{enumerate}
}

To circumvent the hardness of exact (sound and complete) verification, we try to get verifiers with \emph{soundness}, but \emph{not completeness}. That is, an output of \textsc{No Bias} by the verifier $V$ will always be \emph{correct} ($f$ will actually be unbiased). But $\mathcal{V}$ \emph{may} fail to terminate within finite time with the correct output \textsc{No Bias} or a bias instance, even when the classifier is actually unbiased or a valid bias instance exists, respectively. It may also keep finding only spurious bias instances, which we \emph{do not} output.

For developing our algorithms, we formulate the individual bias verification problem as a (non-convex) optimization problem, and use provably-correct global optimization approaches (such as mixed integer linear programming) to perform the verification.

The rationale behind this approach is as follows: If we wish to find a bias instance $x, x^\prime$, the required ``closeness'' constraints $|x_i - x^\prime_i| \leq \varepsilon_j$ (for all $i \in S_j$, for all $j \in [t]$) on $x$ and $x^\prime$ are \emph{linear constraints} (since $|z| \leq \varepsilon \iff -\varepsilon \leq z \leq \varepsilon$), which are easy to handle in an optimization framework (perhaps the only easier form of constraint is an interval constraint $z \in [\alpha, \beta]$). The domain constraints on each $x_i$ are of the form $x_i, x^\prime_i \in [l_i, u_i] \cap (\R \mbox{ or } \Z)$. A problem is that the integrality constraint for discrete attributes is computationally expensive, but this can be mitigated to some extent by the judicious use of relaxations.

We can then say that the model $f$ is individually biased if and only if an input pair $x^\ast, x^{\prime\ast}$ with $|f(x^\ast) - f(x^{\prime\ast})| > \delta$ belongs to the set of pairs $x, x^\prime$ constrained as above. In fact, since $x$ and $x^\prime$ are interchangeable, we can assume $f(x^\ast) - f(x^{\prime\ast}) < -\delta$ for a bias instance. With all these considerations, we can formulate an optimization problem for individual bias verification. 

\begin{equation}\label{prm::ind_bias_optimization}
\begin{aligned}
D^\ast := \min\,\,&f(x) - f(x^\prime)\\
\mbox{s.t. } |x_i - x^\prime_i| &\leq \varepsilon_j,\,\,\forall i \in S_j,\,\,\forall j \in [t]\\
x_i, x^\prime_i \in [l_i, u_i] &\cap (\R \mbox{ or } \Z),\,\,\forall i \in [n] 
\end{aligned}
\end{equation}

If the verifier solves the optimization problem (\ref{prm::ind_bias_optimization}) and finds a solution $(x^\ast, x^{\prime\ast})$ with objective function value $D^\ast < -\delta$, then that solution will be a valid individual bias instance that the verifier can output. On the other hand, if the verifier finds a \emph{certifiable lower bound} which implies $D^\ast \geq -\delta$, then the verifier can correctly output \textsc{No Bias}. However, the requirement of certifiable lower bounds precludes the use of many common optimization techniques based on gradient descent or interpolation.

%For models where the decision function $f$ can be encoded as an SMT formula with linear arithmetic theories, the individual bias verification problem (\ref{prm::ind_bias_optimization}) itself can be encoded as an SMT formula:
%\begin{equation}\label{prm::ind_bias_optimization_smt}
%\begin{aligned}
%y - y^\prime &> \delta,y = f(x), y^\prime = f(x^\prime)\\
%\max(x_i - x^\prime_i,\,&x^\prime_i - x_i) \leq \varepsilon_j\,\,\forall i \in S_j\,\,\forall j \in [t]\\
%x_i, x^\prime_i \in [l_i, u_i] &\cap (\R \mbox{ or } \Z)\,\,\forall i \in [n] 
%\end{aligned}
%\end{equation}

%If (\ref{prm::ind_bias_optimization_smt}) is satisfiable, the model $f$ is individually biased and the satisfiable assignment gives a bias instance. On the other hand, if (\ref{prm::ind_bias_optimization_smt}) is unsatisfiable, it implies that $f$ is unbiased. 

\paragraph{Relaxation}  To avoid the hardness of the integral constraints, it is possible to relax \emph{some} of the categorical features to allow fractional values. This makes sense especially for features such as \emph{age}; which take a large ordered set of values, and where a fractional value is ``interpretable''. It is easy to see that this relaxation preserves the \textsc{No Bias} certification; that is, a classification model is actually unbiased w.r.t. the original attribute domains if it is unbiased w.r.t. the relaxed domains, since we are only expanding the set of valid inputs for bias instances. But this approach may generate spurious counterexamples which we can only reject, and then continue to try to find valid counterexamples. In addition to the domain relaxations, we may also choose to use relaxations of the optimization problem that end up yielding non-tight lower bounds (without valid counterexamples). In general, we may end up doing this indefinitely without actually finding a valid counterexample (even if it exists) --- thus, a verifier that uses a relaxation will not be complete. The specific relaxations that we use in each case will be discussed in later sections.

%A possibility which we do not exploit, is to find approximations for $f$ which are ``good'' (close to $f$, as in $\|f - f_{\rm approx}\|_\infty \rightarrow 0$) as well as ``simple'' (for example, piece-wise linear): in particular a lower-approximation, i.e. an $f_{l}$ such that $f_{l}(x) \leq f(x)$ for all valid $x$, and an upper-approximation, i.e. an $f_{u}$ such that $f_{u}(x) \geq f(x)$ for all valid $x$. Then, we can replace $f$ with the approximations in the constraint solver/optimizer (taking advantage of the fact that $f_u(x) - f_l(x^\prime) \geq f(x) - f(x^\prime)$) to get a sound but incomplete verifier for bias. This is a promising approach, but for the fact that it is not feasible to get both a good lower-approximation as well as a good upper-approximation simultaneously for most functions (linear functions being the exception). For example, it is easy to find a good piece-wise linear lower-approximation for a convex function (using variations of the cutting plane method, for example \parencite{Magnani2009}), but it is not known how to find a good upper-approximation.

The key idea of our work is to demonstrate that we can use the optimization approach described above (with appropriate relaxations and solution methods) to solve the individual bias verification problem for some interesting and useful classes of models, under reasonable assumptions. The details of this, as well as specific techniques and methods, will be given in subsequent sections for each type of classifier that we consider, i.e. linear, kernelized polynomial, and RBF.  We first give a general meta-algorithm for the individual bias verification problem (see Algorithm \ref{alg:meta_ibv_algo}).

The intuition behind this meta-algorithm is just combining the optimization problem formulation (\ref{prm::ind_bias_optimization}) and domain relaxations. We allow the user to choose a \emph{subset} $D$ of the discrete attributes, which should take small sets of values (e.g. boolean attributes). The optimization procedure will fix $x_D$ and $x^\prime_D$ to specific value combinations using equality constraints, and repeat for all such feasible value combinations. The discrete attributes which are not in $D$ \emph{may be (not necessarily)} relaxed to take fractional values. In this way, we can find a set of lower bounds $\ell$ for $f(x) - f(x^\prime)$ which we examine to see if there is any possibility of bias ($\ell < -\delta$). The details of the optimization procedure will vary depending on the type of $f$ (no optimization procedure can solve such a problem in a certifiably optimal way for general $f$), and will be described in later sections.

Note that if no domain relaxations are used \emph{and} the optimization procedure is guaranteed to always find tight bounds (e.g. mixed integer linear programming), then the resulting verifier will be sound and complete. Otherwise, the resulting verifier will be sound but incomplete (e.g. sum-of-squares).

%For the very general function classes found in mathematics --- even the ``nice'' ones such as smooth functions, algebraic functions or square-integrable functions --- there is no computational procedure that can find such lower bounds/certificates. But we show that we can still get useful lower bounds (may be loose bounds, due to relaxations) for certain well-structured function classes of machine learning models.

\begin{algorithm}[t]
\caption{A meta-algorithm for individual bias verification.\label{alg:meta_ibv_algo}}
\begin{algorithmic}[1]
   \Procedure{Verify-Individual-Bias}{$f$, $(S_j)_{[t]}$ $(\varepsilon_j)_{[t]}$, $\delta$, $D$, $(\mathsf{Dom}_i)_{[n]}$}
   \InputState{Classification model $f$ (white-box), discrete attributes $D$, feature partitioning $S_1, \ldots, S_t$, thresholds $\varepsilon_1, \ldots, \varepsilon_t$ and $\delta$, attribute domains $\mathsf{Dom}_1, \ldots, \mathsf{Dom}_n$.}
   \OutputState{Either (i) Valid bias instance $(x, x^\prime)$ or (ii) \textsc{No Bias}.}
   \State{Let $L = \emptyset$.}
   \State{Construct the set $V_p$:
   \begin{equation*}
   \begin{aligned}
   V_p := \{(v,\,&v^\prime) \mid v, v^\prime \mbox{ are feasible for } x_D,\,x^\prime_D\\ \mbox{ and } &|v_i - v^\prime_i| \leq \varepsilon_j \,\,\forall i \in D \cap S_j \,\,\forall j \in [t]\}
   \end{aligned}
   \end{equation*}
   }
      \ForAll{$(v, v^\prime) \in V_p$}
        \State{Let:
		\begin{equation*}
		\begin{aligned}
		D^\ast := \min\,\,&f(x) - f(x^\prime)\\
		\mbox{s.t. } |x_i - x^\prime_i| &\leq \varepsilon_j,\,\,\forall i \in S_j \cap \overline{D},\,\,\forall j \in [t]\\
		x_i, x^\prime_i &\in [l_i, u_i],\,\,\forall i \not\in D\\
		x_D = v &\mbox{ and } x^\prime_D = v^\prime
		\end{aligned}
		\end{equation*}
	    }
       \State{Find a lower bound $\ell \leq D^\ast$.}
       \MultiState{If $\ell < -\delta$, try to find a certificate $x^\ast, x^{\prime\ast}$  for the lower bound $\ell$ (i.e. $f(x^\ast) - f(x^{\prime\ast}) = \ell$), which may not always exist.}
       \MultiState{Add $(\ell, x^\ast, x^{\prime\ast})$ to $L$.}
      \EndFor

   \If{$\ell \geq -\delta$ for all lower bounds in $L$}
      \State{Output \textsc{No Bias}.} 
   \ElsIf{There exists a lower bound $\ell < -\delta$ with a valid certificate in $L$}
       \State{Output bias instance $(x^\ast, x^{\prime\ast})$.}
%   \ElsIf{the time budget has been exceeded, an $L_{ij} < 0$ and no valid certificate}
%       \State{Output \textsc{Possible Bias}.}
   \EndIf
   \EndProcedure
\end{algorithmic}
\end{algorithm}
%\paragraph{Guarantees}  Our verification algorithms are based on Algorithm \ref{alg:meta_ibv_algo} and consider various function classes where we can find the lower bounds (or $\varepsilon$-approximations of lower bounds) computationally. For a sound verifier to terminate, it suffices to find either a certificate that there is no bias instance w.r.t. the relaxed domains or a valid (w.r.t. the original domain constraints) bias instance. However, in general, the relaxations that we use may imply that there could be cases where the classifier is biased w.r.t. the relaxed constraints (and hence a no-bias certificate cannot be found), but is actually unbiased w.r.t. the non-relaxed constraints. Then the verification algorithm will never be able to find a valid bias instance, which leads to \emph{incompleteness} of the verifier. This turns out to not be a problem at all for linear classifiers, where we can get a sound and complete worst-case exponential-time verifier which is quite scalable in practice. Exact verification is possible for piece-wise linear classifiers as well, but in a far less scalable fashion. For kernelized polynomial and RBF classifiers, we get sound but incomplete verifiers. In practice, we set a time budget for the verification algorithms, and terminate with \textsc{Possible Bias/Unknown} output if the verifier is not able to find either a lower bound $\geq -\delta$ or a valid bias instance within that time.
\subsection{LINEAR AND POLYNOMIAL MODELS}\label{sec:poly}

\subsubsection{Linear Models}
We first give an elementary instantiation of the meta-algorithm for linear models, $f(x) = w^\top x + b$. This is the only case where the optimization problem involved is actually convex, and thus the relaxed problem (with continuous attributes) can be solved with soundness and completeness in polynomial-time. In all other cases, the optimization problem is non-convex.

If $f$ is a linear (affine) regression model of the form $f(x) = w^\top x + b$, the objective function of the problem (\ref{prm::ind_bias_optimization}), $f(x) - f(x^\prime) = w^\top x - w^\top x^\prime$, is linear. The constraints are also linear ($|x_i - x^\prime_i| \leq \varepsilon_j$ can be replaced by the pair of linear constraints $x_i - x^\prime_i \leq \varepsilon_j,\, x^\prime_i - x_i \leq \varepsilon_j$), with integrality constraints for the categorical features. Hence the problem (\ref{prm::ind_bias_optimization}) can be solved as a mixed-integer linear program (MILP). MILP solvers can solve the problem \emph{exactly}, modulo computational issues, with worst-case exponential time, and are also fairly efficient in practice for reasonable problem dimensions ($= 2n$ in this case). Thus, we get a sound and complete verifier.

Suppose $f(x) = \sgn(g(x))$, $g(x) = w^\top x + b$, is a linear classification model. Now, the objective function is no longer linear, but since we use $\delta = 0$ for classification, we can take advantage of the fact that $f(x) - f(x^\prime) = 0 \iff g(x) \cdot g(x^\prime) \geq 0$ to get a quadratic objective function for minimization. Since $g(x) \cdot g(x^\prime) = x^\top w w^\top x^\prime + b \cdot (w^\top x + w^\top x^\prime) + b^2$, we can rewrite it in the form $g(x) \cdot g(x^\prime) = (x\,\,x^\prime)^\top Q (x\,\,x^\prime) + b \cdot (w\,\,w)^\top (x\,\,x^\prime) + b^2$, where $Q$ is a positive semi-definite quadratic form in $2n$ variables $(x\,\,x^\prime)$. Hence the problem can be solved \emph{exactly} by mixed-integer quadratic programming (MIQP) solvers, again practically efficient but with worst-case exponential time, to yield a sound and complete verifier.
%\begin{equation*}
%\begin{aligned}
%g(x) \cdot g(x^\prime) &= \begin{bmatrix}x & x^\prime\end{bmatrix} \begin{bmatrix}w w^\top & 0\\0 & 0
%\end{bmatrix} \begin{bmatrix}0 & I_n\\I_n & 0
%\end{bmatrix} %\begin{bmatrix}x\\x^\prime\end{bmatrix}\\
%& + b \cdot \begin{bmatrix}w & w\end{bmatrix} \begin{bmatrix}x\\x^\prime\end{bmatrix} + b^2
%\end{aligned}
%\end{equation*}

\subsubsection{Kernelized Polynomial Models}
In a kernelized classification/regression model with a \emph{polynomial kernel}, the kernel function used is of the form $K(x, y) = (a \, x^\top y + b)^d$, where $a, b \in \R$ are constants and $d$ is the degree of the polynomial kernel. Then,
\begin{equation*}
\begin{aligned}
f(x) &= \sum_{i = 1}^{M} w_i y_i \, K(x_i, x) = \sum_{i = 1}^{M} w_i y_i\,(a\,x_i^\top x + b)^d
\end{aligned}
\end{equation*}

That is, the model $f$ can be viewed as a degree-$d$ polynomial in the variables $x_1, \ldots, x_n$. Thus, the function we wish to lower bound, $f(x) - f(x^\prime)$, is also a degree-$d$ polynomial in $2n$ variables. Let $g(x, x^\prime) := f(x) - f(x^\prime)$. Minimizing $g$ over a linear constraint set ($l_i \leq x_i, x^\prime_i \leq u_i$ for all $i$, $x_i = v_i, x^\prime_i = v^\prime_i$ for all $i \in D$, $-\varepsilon_j \leq x_i - x^\prime_i \leq \varepsilon_j$ for all $i \in S_j$ and for all $j$), as in Algorithm \ref{alg:meta_ibv_algo}, is a polynomial optimization problem over a basic closed semi-algebraic set. If we have discrete attributes which we do not fix during optimization (i.e. not in $D$), but which we do not want to relax, this can be done using polynomial constraints as well. A constraint $x(x_i - 1) \cdots (x_i - k) = 0$ would ensure that $x_i$ takes values in $\{0,1,\ldots,k\}$. The drawback is that such constraints will be very expensive computationally unless $k$ is very small (e.g. $k = 2$ for boolean values).

We can find lower bounds for such polynomial optimization problems using various methods, including sum-of-squares relaxations, geometric programming etc. In this paper, we consider the method of finding lower bounds for polynomial optimization problems using sum-of-squares relaxations (proposed independently by Lasserre and Parrilo, based on the earlier work of Shor \parencite{Shor1987}). We use the particular semidefinite programming (SDP) relaxation from \parencite{LasserreBook} to solve our optimization problem. To the best of our knowledge, this is the first work where sum-of-squares is applied to verify global robustness properties for ML models.

We now give some intuition about the sum-of-squares method for polynomial optimization. The problem is to minimize the given polynomial function $g$ subject to polynomial inequality constraints. The s.o.s algorithm in this case tries to find the largest real number $\gamma$ such that the shifted polynomial $g - \gamma$ can be written as a specific type of polynomial (a polynomial in the \emph{quadratic module} \parencite{LasserreBook} generated by the constraint polynomials). Finding this polynomial can be thought of as finding a vector of monomial coefficients which satisfies appropriate semi-definite constraints. To keep the optimization problem finite-dimensional (i.e. the number of monomials is finite), we have to put an upper-bound $d$ on the degree of this polynomial. This results in the degree-$d$ sum of squares relaxation. With some additional assumptions (always satisfied in our setting), a non-trivial theorem in real algebraic geometry (Putinar's Positivstellensatz \parencite{LasserreBook}) then guarantees that $\gamma$ is a lower bound for the polynomial $g$ on the feasible set.

Suppose $g^\ast$ is the actual tight lower bound for $g$ (i.e. it is achieved by some point which is our certificate/minimizer). However, when the chosen relaxation degree $d$ is too small, we might not be able to represent $g - g^\ast$ as a polynomial of the required form, and the best degree $\leq d$ representation obtained by the optimizer will give a worse (smaller) lower bound. In this case, the verification algorithm must increase $d$ successively to find better lower bounds. It is known that for all such polynomial optimization problems $\mathcal{P}$ and any $\varepsilon > 0$, there will be some finite $d_{\mathcal{P},\varepsilon}$ such that the degree-$d_{\mathcal{P},\varepsilon}$ relaxation lower bound will be $\varepsilon$-close to the actual lower bound (convergence of s.o.s), but there are no known upper bounds for $d_{\mathcal{P},\varepsilon}$ (to the best of our knowledge). Hence an s.o.s-based verification algorithm remains incomplete (unlike the MILP/MIQP approaches), even in cases where all integrality constraints are applied (no domain relaxations), due to this non-zero gap in the s.o.s lower bounds.

%An \emph{extremely simplified} depiction of the process (with univariate polynomials of increasing degree) is shown in Figure \ref{fig::poly::func_approx}. The actual polynomials involved cannot be depicted graphically, even in the simplest case of degree-$2$ polynomial kernels, since they are $n$-variate  ($n$ = number of features $\gg 2$) and have degree $\geq 4$. 

%\begin{figure}[t]
%    \centering
%    \includegraphics[width=0.9\columnwidth]{images/function_approx_graph.png}
%    \caption{Sum-of-squares intuition: Successive approximations by polynomial functions of higher degree. The red line shows the actual polynomial function $f$ and the three lines below show sum-of-squares approximations of increasing degrees.}
%    \label{fig::poly::func_approx}
%\end{figure}

We omit a full description of the sum-of-squares relaxations, but the book by Lasserre \parencite{LasserreBook} is a good reference which includes all the details and proofs. A brief technical description of the relaxations (\textit{sans} proofs) is given in the supplementary material.

Using the sum-of-squares relaxations, and the fact that semidefinite programs can be solved up to exponential accuracy in polynomial time (w.r.t the number of variables and constraints of the SDP), we get the following:

\begin{theorem}\label{thm:poly_opt_sos}
There is a polynomial-time algorithm (which runs in time $n^{O(2d)}$) that outputs $\ell \pm O(1/2^{n})$, where $\ell$ is a lower bound for $g(x, x^\prime) := f(x) - f(x^\prime)$, subject to a linear constraint set as in Algorithm \ref{alg:meta_ibv_algo}.
\end{theorem}

The $\pm O(1/2^n)$ error comes from the error in solving semi-definite programs, which cannot be avoided even when using exact arithmetic (since SDPs with rational coefficients need not have rational solutions).

This implies that we have a sound but incomplete algorithm to solve the relaxed individual bias verification problem for polynomial kernel classifiers, by plugging in the sum-of-squares relaxation algorithm from Theorem \ref{thm:poly_opt_sos} into the meta-algorithm (Algorithm \ref{alg:meta_ibv_algo}) from Section \ref{sec:meta_algo}.

The salient points of our technique for polynomial kernelized classifiers are give below:

\begin{itemize}
\denseitems
    \item We use the sum-of-squares relaxation technique to find a lower-bound approximation of $g(x, x^\prime)$.
    \item The lower-bound approximation ensures that when our verifier is always correct when it says \textsc{No Bias}. However, it can yield spurious counter-examples / loose lower bounds at any particular level (of relaxation degree).
    \item The approximations are refined by taking sum-of-squares relaxations of higher degree, which makes the s.o.s lower bound closer to the actual lower bound of $g(x, x^\prime)$.
    \item This refinement process may not terminate (only a convergence result is known), yielding an incomplete verifier.
\end{itemize}

\subsection{RBF Kernelized Classifiers}\label{sec:rbf}

The Radial Basis Function (RBF) kernel is of the form $K(x, y) = \exp(-\gamma \|x - y\|_2^2)$, for a fixed parameter $\gamma$. So the kernelized classifier is of the form
\begin{equation*}
\begin{aligned}
    f(x) = \sum_{i \in \mathcal{S}^+} \varphi_i(x) - \sum_{i \in \mathcal{S}^-} \varphi_i(x)
\end{aligned}
\end{equation*}
where $\varphi_i(x) := w_i\,\exp(-\gamma \|x - x_i\|_2^2)$. We use $\mathcal{S}^+$ to denote the subset of indices $i$ with $y_i = 1$, and $\mathcal{S}^-$ to denote those with $y_i = -1$. We will abuse the notation to write $x_i \in \mathcal{S}^+$ and $i \in \mathcal{S}^+$ as appropriate, and similarly for $\mathcal{S}^-$. Suppose that all the non-zero model weights satisfy $0 < c < w_i < C$ for some bounds $c$ and $C$. Let $g(x, x^\prime) := f(x) - f(x^\prime)$.

Let $\epsilon > 0$ be a very small constant (compared to $c$). Suppose that, when searching for bias instances, we only want to find pairs $x, x^\prime$ where $g(x, x^\prime) < -2\varepsilon$. Then, we can completely avoid looking at regions with $|f(x)| \leq \varepsilon$ and $|f(x^\prime)| \leq \varepsilon$ (by triangle inequality).  We can directly exploit the above fact when $\delta > 0$ (with $\varepsilon = \delta/2$) since our desired condition for bias is $g(x, x^\prime) < -\delta$. Things are not as straightforward when $\delta = 0$ (with classification models), but we argue that, under reasonable assumptions, we can fix a sufficiently small (but non-zero) $\varepsilon$, say $\varepsilon = 10^{-8}$, such that finding a lower bound $g(x, x^\prime) \geq -2\varepsilon$ rather than $g(x, x^\prime) \geq 0$ does not exclude any valid and \emph{interesting} bias instances. Specifically, the assumption is that we consider the non-zero model weights ($w_i > c$) to not be too small compared to $\varepsilon$, and we are not interested in bias instances where the attribute values have small positive magnitude of the order of $\varepsilon$ (such attribute values are unlikely to occur in real world data), and based on the bounds that we specify, we are not considering points $x$ which are very far from all the support vectors. Hence we can argue that, for a very small $\varepsilon > 0$,  we have $|f(x)| \geq \varepsilon$ for all valid $x$ which we wish to consider as a bias instance.

\begin{theorem}
Given a kernelized classifier (with the RBF kernel) $f: \R^n \rightarrow \R$ with $M$ support vectors, and for a fixed, sufficiently small, $\varepsilon > 0$, there is an algorithm \emph{\textsc{Find-Bias-Rbf}} which runs in time $\poly(n, M)$ that either (i) returns a bias instance $x, x^\prime$ with $g(x, x^\prime) := f(x) - f(x^\prime) < -2\varepsilon$, or (ii) returns a lower-bound $L \geq -2\varepsilon$ such that, for all valid data points \footnote{Assuming that we do not have valid feature values as small as $\approx n \sqrt{\varepsilon}$.}, we have $g(x, x^\prime) \geq L$.
\end{theorem}

We mention that we chose $\varepsilon = 10^{-8}$ in our experiments based on the above considerations, because the datasets we used had $\approx 10-20$ features, and we decided (rather unilaterally) that bias instances with feature values $\ll 10^{-3}$ were not interesting.

We now give a sketch of both the algorithm and the proof of correctness. A detailed pseudocode is left for the supplementary material. We note that this algorithm (\textsc{Find-Bias-Rbf}) is slightly different compared to the meta-algorithm in Section \ref{sec:meta_algo}. However, it can still be plugged into the meta-algorithm to verify individual bias of a given kernelized RBF classifier. The main differences are: (i) {In the case of finding a valid bias instance $(x, x^\prime)$, the algorithm \textsc{Find-Bias-Rbf} does not give an exact lower bound $\ell$ as in the meta-algorithm (we can only say that $\ell < -2\varepsilon$)}, and (ii) {The case where the \textsc{Find-Bias-Rbf} algorithm gives a lower bound $\geq -2\varepsilon$ is treated as the \textsc{No Bias} case (which differs from the meta-algorithm when $\delta = 0$).} But we still get a correct output in case (i), and have already justified that a very small $\varepsilon$ will make sure that all valid examples are covered in case (ii); with only minor assumptions on the minimum magnitudes of the model weights and the attribute values.

The intuition behind the \textsc{Find-Bias-Rbf} algorithm is as follows. A kernelized RBF classifier $f$ is a linear combination of $n$-dimensional gaussian densities, where for each support vector $x_i$ (with label $y_i$), you have a gaussian with mean (i.e. centered at) $x_i$ with covariance matrix $\tfrac{1}{2\gamma}I$. This gaussian is scaled by a factor $0 < w_i < C$, which is equivalent to scaling the variance of each (i.i.d.) co-ordinate by $w_i^2 < C^2$. Then, we can write
\begin{equation*}
\varphi_i(x) = \Norm{x_i,\tfrac{w_i^2}{2\gamma} I}(x)
\end{equation*}
where we again abuse the notation and use $\Norm{\mu,\Sigma}$ to denote the gaussian probability density function (pdf) with mean $\mu$ and covariance matrix $\Sigma$.

Now, for any $x, x^\prime$ such that $f(x) < -\varepsilon$ and $f(x^\prime) > \varepsilon$ (which would imply that $g(x, x^\prime) < -2\varepsilon$), there must exist a support vector $x_r \in \mathcal{S}^+$ such that $\varphi_r(x^\prime) \geq \varepsilon / M$, and a support vector $x_s \in \mathcal{S}^-$ such that $\varphi_s(x) \geq \varepsilon / M$. That is, any individual bias instance (pair of inputs) that we care about (w.r.t. our fixed $\varepsilon$) must be in the intersection of $\mathcal{B}_{\ell_2}\left(x_r, D_r\right)$ and $\mathcal{B}_{\ell_2}\left(x_s, D_s\right)$ for some support vectors $x_r \in \mathcal{S}^+$ and $x_s \in \mathcal{S}^-$, where $D_r := \sqrt{\tfrac{1}{\gamma}\log\left(\tfrac{M w_r}{\varepsilon}\right)}$ (and $D_s$ is defined similarly), which means that it must also be in the intersection of $\mathcal{B}_{\ell_\infty}\left(x_r, D_r\right)$ and $\mathcal{B}_{\ell_\infty}\left(x_s, D_s\right)$. We then minimize the objective function $P(x, x^\prime) = \tfrac{1}{2}\big(\sum_{u \in \mathcal{S}^+} w_u \|x^\prime - x_u\|^2$ $+ \sum_{v \in \mathcal{S}^-} w_v \|x - x_v\|^2\big)$, subject to the linear constraints that $x, x^\prime \in \mathcal{B}_{\ell_\infty}(x_r, D_r) \cap \mathcal{B}_{\ell_\infty}(x_s, D_s)$.

It is easy to see that minimizing $P(x, x^\prime)$ subject to these constraints (for a particular pair $x_r$ and $x_s$) is a convex quadratic program, which can be solved in $\poly(n)$ time. This has to be done for all $x_r$ and $x_s$. This requires $\leq M^2$ iterations, and so the entire algorithm runs in $\poly(n, M)$ time. If no appropriate bias instance is found in these iterations, we output the lower bound $L \geq -2\varepsilon$, which is the smallest value of $g(x^\ast, x^{\prime\ast})$ among those found in each iteration.

\section{EXPERIMENTAL RESULTS}\label{sec:experiments}

\begin{table}[t]
\scriptsize
	\centering \renewcommand{\arraystretch}{1.2}
	\caption{Experimental results of the proposed algorithms for the relaxed problem.\label{tbl:exp_res1}}
	\begin{tabular}{llrrcr}
		\toprule
		\multirow{2}{*}{\textbf{DS}} &
		\multirow{2}{*}{\textbf{Model}} &
		\multicolumn{2}{c}{\textbf{Accuracy}} &
		\multirow{2}{*}{\textbf{Bias}} &
		\multirow{2}{*}{\textbf{Time taken}}\\
		\cmidrule(lr){3-4}
		& & Train & Test & & \\
		\midrule
		\multirow{6}{*}{GC}
		         & \textsf{GC\_Linear1} & 0.767 & 0.76 & Yes & 226.2s\\
	             & \textsf{GC\_Linear2} & 0.756 & 0.748 & No  & 229.0s\\
		         & \textsf{GC\_Poly1}   & 0.650 & 0.704 & No & 78.4m\\
		         & \textsf{GC\_Poly2}   & 0.658 & 0.7 & No  & 77.3m\\
		         & \textsf{GC\_Rbf1}    & 0.992 & 0.664 & Yes & 9.5s\\
		         & \textsf{GC\_Rbf2}    & 1.0 & 0.7 & No  & 27.7m\\
		\midrule
		\multirow{2}{*}{AD}
		         & \textsf{AD\_Linear1} & 0.822 & 0.821 & Yes & 47.3s\\
		         & \textsf{AD\_Linear2} & 0.823 & 0.821 & No  & 48.7s\\
		         & \textsf{AD\_Poly1}    & 0.826 & 0.824 & Possible & 10.4s\\
		         & \textsf{AD\_Poly2}    & 0.826 & 0.823 & Possible & 10.9s\\
		         & \textsf{AD\_Rbf1}    & 0.828 & 0.827 & Yes & 114.5s\\
		         & \textsf{AD\_Rbf2}    & 0.905 & 0.811 & Yes  & 546.8s\\
		\midrule
		\multirow{6}{*}{FD}
		         & \textsf{FD\_Linear1} & 0.660 & 0.654 & Yes & 0.163s\\	
		         & \textsf{FD\_Linear2} & 0.660 & 0.654 & No  & 0.101s\\
		         & \textsf{FD\_Poly1}   & 0.602 & 0.610 & Possible & 18.1s\\	
		         & \textsf{FD\_Poly2}   & 0.590 & 0.607 & Possible  & 17.1s\\
		         & \textsf{FD\_Rbf1}    & 0.928 & 0.669 & Yes & 16.1s\\	
		         & \textsf{FD\_Rbf2}    & 1.0 & 0.665 & Yes  & 32.7s\\
		\midrule
		\multirow{6}{*}{CR}
		         & \textsf{CR\_Linear1} & 0.943 & 0.86 & Yes & 106.8s\\	
		         & \textsf{CR\_Linear2} & 0.953 & 0.88 & No  & 79.1s\\
		         & \textsf{CR\_Poly1}   & 0.906 & 0.93 & No Bias & 33.8s\\
		         & \textsf{CR\_Poly2}   & 0.863 & 0.9 & No Bias  & 31.8s\\
		         & \textsf{CR\_Rbf1}    & 1.0 & 0.85 & Yes & 9.2s\\	
		         & \textsf{CR\_Rbf2}    & 1.0 & 0.56 & No  & 389.4s\\
		\bottomrule
	\end{tabular}
\end{table}

\begin{table}[t]
\scriptsize
    \centering
    \renewcommand{\arraystretch}{1.2}
    \caption{Experimental results of testing bias using random sampling.\label{tbl:exp_res2}}
    \begin{tabular}{lrl}
        \toprule
        \textbf{Model}     & \textbf{Time Taken} & \textbf{Result}\\
        \midrule
        GC\_Lin1   & 1.812 mins & Not found\\
        GC\_Lin2   & 2.088 mins & Not found\\
        GC\_Poly1  & 1.756 mins & Not found\\
        GC\_Poly2  & 1.646 mins & Not found\\
        GC\_Rbf1   & 4.039 mins & Not found\\
        GC\_Rbf2   & 4.737 mins & Not found\\
        \bottomrule
    \end{tabular}
\end{table}

\subsection{SETUP}
All our experiments are carried out on a cloud virtual machine with $32$ Intel Xeon E5-2683 v4 (2.10 GHz) processors, $128$ GB RAM and no dedicated GPU. The machine runs Ubuntu 16.04, and has Python 3.6.8 (Anaconda) installed, along with all the default Anaconda packages. Each experiment is run with at most $32$ parallel jobs (\texttt{python} processes) on this machine.

\textbf{Tools} We use the \texttt{cplex} and \texttt{quadprog} Python packages to solve the quadratic programs, and \texttt{SDPA} to solve the sum-of-squares relaxation SDPs.

\begin{table}[t]
    \centering
    \caption{The datasets used for experiments\label{tbl:datasets}}
    \begin{tabular}{lcc}
         \toprule
         Dataset & \# Features & \# Rows (Train)\\
         \midrule
         German Credit (GC) & 20 & 750\\
         Adult (AD) & 8 & 24420\\
         Fraud Detection (FD) & 9 & 825\\
         Credit ISLR (CR) & 10 & 300\\
         \bottomrule
    \end{tabular}
\end{table}

\subsection{BENCHMARKS}

\subsubsection{Datasets}
We use four publicly available datasets to benchmark our algorithms, listed in Table \ref{tbl:datasets}. Note that the Adult dataset which we use is a publicly available dimension-reduced variant \parencite{DS_AD} of the full dataset. In all cases, we report the time taken for the individual bias verification step on an already trained model. The times which we report are obtained from the Python \texttt{time.perf\_counter()} function.

In all the datasets, we replace textual categorical attributes by appropriate integer values as appropriate. We also we do a $75\%-25\%$ stratified split on each dataset before using it for training.

For the perturbation bounds, we use a counterfactual formulation where a subset of the attributes in each dataset are selected as protected/sensitive, with arbitrary perturbations allowed ($\varepsilon_1 = \infty$), and the rest of the attributes are fixed ($\varepsilon_2 = 0$). The protected attributes are: \textrm{sex-marital-status} for German Credit, \textrm{race} for Adult, \textrm{ethnicity} for Fraud Detection, and \textrm{gender, ethnicity} for Credit. 

\subsubsection{Models} All the models are trained using the \texttt{scikit-learn} framework. The linear models are trained using scikit\-learn logistic regression, with $L_2$ regularization and the default parameters. The \textsf{DD\_Linear2} models are trained after setting the protected attribute values to $0$ throughout the training data (masking). The rbf kernelized models are trained using the support vector machine classifier (\texttt{sklearn.svm.SVC}) with the rbf kernel. The \textsf{DD\_Rbf1} models are trained with $C = 1000$ and $\gamma = 10^{-4}$. The \textsf{DD\_Rbf2} models are trained with $C = 1$ and $\gamma = 0.5$, and after masking the protected attribute. The polynomial kernelized models are trained using \texttt{SVC} with the degree-$2$ polynomial kernel. The \textsf{DD\_Poly1} models are trained with $C = 1.$, $\gamma = 0.001$, and $r = 0$. The \textsf{DD\_Poly2} models are trained with the same hyperparameters, but after masking the protected attribute.

\subsection{EVALUATION OF VERIFICATION}
We show the results of evaluating our proposed verification algorithms on the models described above and show the results in Table \ref{tbl:exp_res1}. It can be seen that the time taken for verification --- even for reasonably complex models on real-world datasets --- is within fairly acceptable limits, even with the worst case exponential time. It can also be seen that the sum-of-squares relaxations scale quite badly as the dimension of the data increases, which is as expected.

Wherever Table \ref{tbl:exp_res1} shows Bias = Yes, it indicates that the verifier gave a valid bias instance as the output. Bias = No indicates that the verifier proved No Bias. Bias = Possible indicates that the verifier did not prove a lower-bound $\geq 0$, but neither did it find a valid bias instance. An example of a bias instance found for the model \textsf{AD\_Lin1} is (age: 60, work: Private, edu: Bachelors, marital-status: Married, occupation: Professional, race: White, sex: Male, hours-per-wk: 8). The model predicts the income as $\geq 50k$. Changing race from ``White'' to ``Other'' flips the income prediction to $< 50k$.

\subsection{COMPARISON WITH TESTING}
We performed an experiment by running the random testing algorithm (THEMIS) with 50,000 samples and verification algorithm to compare the time taken by the verification algorithm and testing for finding counter-example and determine that our algorithm provides NO BIAS before the testing algorithm exhaust the test case generation in a specific time. The result is presented in Table \ref{tbl:exp_res2}. The result shows that random testing can take comparable or worse time than our verification algorithm, even without generating a single counterexample (bias instance). 

\section{CONCLUSION AND DISCUSSION}\label{sec:conclusion}
We have considered a notion of individual fairness for structured data, and the problem of verifying the lack of individual bias in a given decision model. We have given a meta-algorithm for solving this problem, as well as specific algorithms for linear models and kernelized models with polynomial/RBF kernels. To the best of our knowledge, this is the first work that considers the verification of individual fairness for ML models.

\textbf{Analysis of Model Bias } Our algorithms output either a no-bias certificate or a bias instance (input pair), but this may be insufficient by itself in real-world investigations of model bias. To further the analysis, our solution offers two possibilities out-of-the-box. One possibility is to tighten the constraints on the input features, invoking the verifier every time, to find different \emph{input regions} where the model is fair. Another possibility is to impose additional linear constraints (domain-knowledge-based) on the bias instances, which does not affect the optimization formulations. \emph{Note that we are not reporting any experimental results for the above analyses}. Another important problem is to find regions where bias exists for \emph{every} input in that region, but this is not possible using our verifier alone.

\textbf{Future Work} In future, we plan to extend this work in the following dimensions --- 1) verifying wider classes of ML models, 2) extending our techniques to work with other individual \& group fairness definitions, and 3) exploring different abstraction-refinement schemes such as counter-example driven refinement.

\textbf{Acknowledgements} The authors would like to thank Dinesh Garg and Rishi Saket for helpful discussions.

% \printbibliography
\section*{SUPPLEMENTARY MATERIAL}
% \begin{refsection}
\subsection*{PRELIMINARIES}
We now provide some terminology and notations related to polynomials and polynomial optimization using sum-of-squares. An $n$-variate real polynomial $P$ is a sum of finitely many terms of the form $c_\alpha x_1^{\alpha_1} \cdots x_n^{\alpha_n}$ where $\alpha = (\alpha_1, \ldots, \alpha_n) \in \N^n$ and $c_\alpha \in \R$. The monomial $x_1^{\alpha_1} \cdots x_n^{\alpha_n}$ is also denoted by $x^\alpha$, and the polynomial $P$ can be written as $P(x) = \sum_{\alpha \in \N^n} c_\alpha x^\alpha$ where $c_\alpha \neq 0$ only for finitely many $\alpha$. The degree of a monomial $x^\alpha$ is $|\alpha| := \alpha_1 + \ldots + \alpha_n$, and the degree of a polynomial is the maximum degree of all its monomials with non-zero coefficients.

A $n$-variate polynomial $P = \sum_{\alpha} c_\alpha x^\alpha$ with degree $\leq d$ can be associated with its coefficient vector $(c_\alpha)$ as a point in $\R^{s_n(d)}$, where $s_n(d) := \binom{n + d}{d} = O(n^{d})$ (which can be seen by counting the monomials $x^\alpha$ with $\alpha \in \N^n$ and $|\alpha| = \alpha_1 + \ldots + \alpha_n \leq d$).

A set $K \subseteq \R^n$ is said to be a \emph{(basic closed) semi-algebraic set} if there exist $n$-variate polynomials $g_1, \ldots, g_m$ such that $$K = \left\{x \in \R^n : g_i(x) \geq 0 \mbox{ for all } i \in [m]\right\}.$$

\subsection*{SUM-OF-SQUARES RELAXATIONS}\label{sec:poly:sos_relaxations}

A polynomial $P$ is said to be a sum-of-squares (s.o.s) if there exist some $m \geq 1$ and $n$-variate polynomials $G_1, \ldots, G_m$ such that $P = G_1^2 + \ldots + G_m^2$. The set of polynomials $G := \{G_1, \ldots, G_m\}$ is said to be a \emph{sum-of-squares decomposition} of $P$. The degree of the s.o.s decomposition is defined to be $\deg(G) := \max_{i \in [m]} \deg(G_i)$. A polynomial $P$ is said to be a degree-$d$ sum-of-squares if it has a s.o.s decomposition of degree $\leq d$. Clearly, a polynomial which is degree $d$ s.o.s has degree $\leq 2d$, and the degree of a s.o.s representation for a degree $\leq 2d$ polynomial (if it exists) is at most $d$.

It is easy to see that every s.o.s polynomial is non-negative or positive semi-definite (p.s.d), but the converse (every p.s.d polynomial is s.o.s) is not true except in very specific cases (univariate polynomials, quadratics, bivariate quartics), as proved by Hilbert \parencite{Hilbert1888}.

However, we can construct a sound, but incomplete, verifier for the non-negativity (p.s.d-ness) of a given polynomial by checking whether the polynomial has degree-$d$ sum-of-squares decomposition (for appropriately large $d$). Shor \parencite{Shor1987} showed that the question of whether a given polynomial $f$ has a degree-$d$ sum-of-squares decomposition is equivalent to the feasibility of a semidefinite program (SDP) with $O(n^{2d})$ variables and $O(n^{d})$ constraints. For constant $d$, such an SDP (which we may call the degree-$d$ s.o.s relaxation) can be solved in $\poly(n)$ time.

Let $[x]_d$ denote the $s_n(d)$-length vector of all $n$-variate monomials with degree $\leq d$, according to some monomial ordering. Say,
\begin{equation*}
[x]_d := (1 \,\,\, x_1 \, \cdots \, x_n \,\,\, x_1^2 \,\,\, x_1x_2 \, \cdots \, x_1 x_n \,\,\, \,\,\, x_n^2 \, \cdots \, x_1^d \, \cdots \, x_n^d).
\end{equation*}

Let $f$ be a $n$-variate real polynomial with $\deg(f) \leq 2d$. That is,
\begin{equation*}
f = \sum_{\stackrel{\alpha}{|\alpha| \leq 2d}} c_\alpha x^\alpha = c^\top [x]_{2d}\,\,\,\mbox{ for some } c \in \R^{s_n(2d)}.
\end{equation*}

\begin{theorem}[\parencite{Shor1987}]\label{thm::sos_feasiblility}
	$f$ is degree-$d$ s.o.s if and only if there exists a symmetric positive semidefinite matrix $Q \in \R^{s_n(d) \times s_n(d)}$ such that $f = [x]_d^\top Q [x]_d$, coefficient-wise. That is, $c_\alpha = \sum_{\beta + \gamma = \alpha} Q_{\beta, \gamma}$ for all $\alpha$ such that $x^\alpha \in [x]_{2d}$, and $\beta, \gamma$ such that $x^\beta, x^\gamma \in [x]_d$. 
\end{theorem}

To perform (unconstrained) polynomial optimization --- i.e. finding the global minimum $f^\ast := \inf_{x \in \R^n} f(x)$ of a given polynomial function $f$ --- using sum-of-squares, Shor \parencite{Shor1987} formulated a sequence of sum-of-squares relaxation SDPs (which have increasing size/complexity as the degree $d$ increases). The degree-$d$ SDP finds $f_{\rm sos}^{(d)} := \sup \gamma$, s.t. $f - \gamma$ is a degree-$d$ s.o.s (which implies that $\gamma$ is a lower bound for $f$).

\begin{equation*}
\begin{aligned}
\max_{Z}\,-A^{(\mathbf{0})} &\circ Z,\,\,\,\,\mbox{subject to}\\
A^{(\alpha)} \circ Z = c_\alpha\,\,(\mbox{where } A^{(\alpha)}_{\beta,\gamma} &= 1 \mbox{ if } \beta + \gamma = \alpha \mbox{ and } 0 \mbox{ othewise.})\\
(\mbox{for all } \alpha &\neq \mathbf{0} \in \N^n_d)\\
Z \succeq 0,\,\,&Z \in \mathbb{S}^{s_n(d)}(\R)
\end{aligned}
\end{equation*}

The dual of the above SDP is

\begin{equation*}
\begin{aligned}
\min_{y}\,c^\top y&,\,\,\,\,\mbox{subject to}\\
\sum_{\alpha} y_\alpha \cdot A^{(\alpha)} &\succeq 0,\,\,y_{\mathbf{0}} = 1,\,\,y \in \R^{s_n(2d)}
\end{aligned}
\end{equation*}

In the above SDP, $Z$ may be interpreted as $Z \equiv Q - \gamma \cdot E_{11}$, where $Q \in \mathbb{S}^{s_n(d)}(\R)$ is a symmetric p.s.d matrix such that $[x]_d^\top Q [x]_d = f$ (as in Theorem \ref{thm::sos_feasiblility}), and $E_{11}$ denotes the elementary matrix with a $1$ in the (first row, first column) and zeros elsewhere. 

This implies, of course, that the objective is $-A^{(\mathbf{0})} \circ Z = -Z_{\mathbf{0},\mathbf{0}} = \gamma - c_{\mathbf{0}}$; maximizing it is equivalent to maximizing $\gamma$, and the s.o.s lower bound $f_{\rm sos}^{(d)} := \gamma$ may be recovered as $\gamma = c_{\mathbf{0}} - Z_{\mathbf{0},\mathbf{0}}$.

This hierarchy of SDPs gives a sequence of increasing lower bounds $f_{\rm sos}^{(1)} \leq f_{\rm sos}^{(2)} \leq f_{\rm sos}^{(3)} \leq \ldots$ for $f$, where we define $f_{\rm sos}^{(d)} := -\infty$ if the degree-$d$ SDP is infeasible. It is also possible in some cases (with s.o.s relaxations of sufficiently high degree) to extract a \emph{certificate} $x^\ast$ for the lower bound (i.e. an $x^\ast \in \R^n$ such that $f(x^\ast) = f_{\rm sos}^{(d)}$). Clearly, the existence of such a certificate implies that the sum-of-squares hierarchy has reached the actual optimum, i.e. $f_{\rm sos}^{(d)} = f^\ast$. However, this will not occur for all polynomials (and hence the s.o.s-based non-negativity verifier will always be \emph{incomplete}). In the unconstrained minimization case, we only need to check s.o.s relaxations of degree $\leq 2d$. But such a degree upper-bound is not known for the constrained case, which is described below.

Finally, it was shown by Lasserre \parencite{Lasserre2001} and Parrilo \parencite{Parrilo2000} independently that it is possible to lower-bound a polynomial optimization problem over a basic closed semi-algebraic set $K \subseteq \R^n$ by using semi-definite relaxations --- a (basic closed) semi-algebraic set $K \subseteq \R^n$ is the intersection of the solution sets of finitely many non-strict inequalities of real polynomials. This is done by applying results from real algebraic geometry known as \emph{Positivstellensatz}.

If $\mathbb{K} = \{x \in \R^n : g_i(x) \geq 0 \mbox{ for all } i = 1, \ldots, m\}$ is a \emph{compact} semi-algebraic set, then we can get sum-of-squares lower bounds (using Putinar's Positivstellensatz) for the constrained polynomial optimization problem of finding $f_{\mathbb{K}}^\ast := \inf_{x \in \mathbb{K}} f(x)$ as follows:

Let $v_j := \lceil \deg(g_j)/2 \rceil$, and let $$d \geq d_0 := \max(\lceil \deg(f)/2 \rceil, v_1, \ldots, v_m).$$ Then
\begin{equation*}
\begin{aligned}
f_{\rm sos}^{(d)} &:= \sup \gamma, \mbox{ s.t.}\\
f - \gamma &= \sum_{j = 1}^{m} \sigma_j g_j,\mbox{ where } \sigma_j \mbox{ is degree-}(d - v_j) \mbox{ s.o.s}
\end{aligned}	
\end{equation*}

Similar to the unconstrained case, this gives a sequence of increasing lower bounds for  $f^\ast$.

\begin{algorithm}[t]
	\caption{Individual bias verification for kernelized RBF models.\label{alg:find_bias_rbf}}
	\begin{algorithmic}[1]
		\Procedure{Find-Bias-Rbf}{}
		\MultiState{Let $f = \sum_{i = 1}^{M} w_i y_i \exp(-\gamma\|x - x_i\|^2)$,  with $c < w_i < C$ for all $i \in [M]$.}
		\MultiState{Let $\mathcal{S}^+$ be the subset of $\{(x_i, y_i)\}_{i = 1}^{M}$ with $y_i = 1$ and $\mathcal{S}^-$ be the subset with $y_i = -1$.}
		\State{Let $L := \emptyset$.}
        \State{Construct the set $V_p$:
        \begin{equation*}
        \begin{aligned}
        V_p := \{(v,\,&v^\prime) \mid v, v^\prime \mbox{ are feasible for } x_D,\,x^\prime_D\\ \mbox{ and } &|v_i - v^\prime_i| \leq \varepsilon_j \,\,\forall i \in D \cap S_j \,\,\forall j \in [t]\}
        \end{aligned}
        \end{equation*}
        }
		\ForAll{$(v, v^\prime) \in V_p$}
		\ForAll{$x_r \in \mathcal{S}^+$}
		\ForAll{$x_s \in \mathcal{S}^-$}
			\State{Solve this optimization problem to get $x^\ast, x^{\prime\ast}$:}
			\State{
			\begin{equation*}
			\begin{aligned}
				\min_{\mathrm{valid}\,\,x,\, x^\prime}\,\tfrac{1}{2} \bigg(\sum_{u \in \mathcal{S}^+} w_u &\|x^\prime - x_u\|^2 + \sum_{v \in \mathcal{S}^-} w_v\|x - x_v\|^2\bigg)\\
				&\mbox{subject to}\\
				x_{rk} - D_r &\leq x_k, x^\prime_k \leq x_{rk} + D_r \mbox{ and }\\
				x_{sk} - D_s &\leq x_k, x^\prime_k \leq x_{sk} + D_s, \mbox{ for all } k\\
				|x_i - x^\prime_i| &\leq \varepsilon_j\,\,\,\forall i \in S_j \cap \overline{D} \,\,\forall j \in [t]\\
				x_D = v \mbox{ and } x^\prime_D = v^\prime
			\end{aligned}
			\end{equation*}
			}
			\If{$f(x^{\prime\ast}) \geq \varepsilon$ and $f(x^\ast) \leq -\varepsilon$}
				\State{Output $(x^\ast, x^{\prime\ast})$ and \Return}
			\Else
				\State{Add $f(x^{\prime\ast}) - f(x^\ast)$ to $L$.}
			\EndIf
		\EndFor
		\EndFor
		\EndFor
		\State{Output the lower bound $L^\ast := \min L$.}
		\EndProcedure
	\end{algorithmic}
\end{algorithm}

\printbibliography
% \end{refsection}

\end{document}